\documentclass[journal,twoside,web]{ieeecolor}
\usepackage{amsmath,amssymb,amsfonts}
\usepackage{algorithmic}
\usepackage{array}
\usepackage{cite}
\usepackage{float}
\usepackage{generic}
\usepackage{graphicx}
\usepackage{multirow}
\usepackage{tabularx}
\usepackage{textcomp}
\usepackage{verbatim}

\newcolumntype{L}[1]{>{\raggedright\arraybackslash}p{#1}}
\def\BibTeX{{\rm B\kern-.05em{\sc i\kern-.025em b}\kern-.08em
    T\kern-.1667em\lower.7ex\hbox{E}\kern-.125emX}}
\begin{document}
\title{Two-Stage Hierarchical and Explainable Feature Selection Framework for Dimensionality Reduction in Sleep Staging}
\author{Yangfan Deng, Hamad Albidah, Ahmed Dallal, Jijun Yin, and Zhi-Hong Mao
\thanks{Yangfan Deng, Hamad Albidah, Ahmed Dallal, and Jijun Yin are with the Department of Electrical and Computer Engineering, University of Pittsburgh, Pittsburgh, PA 15261 USA.}
\thanks{Zhi-Hong Mao is with the Department of Electrical and Computer Engineering and the Department of Bioengineering, University of Pittsburgh, Pittsburgh, PA 15261 USA (e-mail: zhm4@pitt.edu).}}

\maketitle

\begin{abstract}
Sleep is crucial for human health, and EEG signals play a significant role in sleep research. Due to the high-dimensional nature of EEG signal data sequences, data visualization and clustering of different sleep stages have been challenges. To address these issues, we propose a two-stage hierarchical and explainable feature selection framework by incorporating a feature selection algorithm to improve the performance of dimensionality reduction. Inspired by topological data analysis, which can analyze the structure of high-dimensional data, we extract topological features from the EEG signals to compensate for the structural information loss that happens in traditional spectro-temporal data analysis. Supported by the topological visualization of the data from different sleep stages and the classification results, the proposed features are proven to be effective supplements to traditional features. Finally, we compare the performances of three dimensionality reduction algorithms: Principal Component Analysis (PCA), t-Distributed Stochastic Neighbor Embedding (t-SNE), and Uniform Manifold Approximation and Projection (UMAP). Among them, t-SNE achieved the highest accuracy of 79.8\%, but considering the overall performance in terms of computational resources and metrics, UMAP is the optimal choice.
\end{abstract}


\section{Introduction}
\label{sec:introduction}
\IEEEPARstart{S}{leep} plays a crucial role in both physiological and psychological health for human beings. Sleep serves as a period of rest for both the body and mind, characterized by a reduction or total absence of consciousness and minimal bodily movements. Several important processes take place during sleep, including recovery and regeneration, as well as the consolidation that transforms short-term memories into long-term memories. The lack of sufficient sleep is concerned with high risk of developing cardiovascular illnesses, metabolic and hormonal imbalances, obesity, and other health issues. The sleep cycle in a typical night consists of 4-6 cycles. Each cycle is segmented into four phases: stages N1-N3 and rapid eye movement (REM)~\cite{brodbeck2012eeg}. The different sleep stages are distinguishable by the brain wave frequencies, indicating the depth and quality of the sleep. In all potential techniques for understanding sleep patterns, the analysis of electroencephalograms (EEGs) stands out for the capability to reveal the sophisticated activities of the brain.

EEG signals are recorded by electrodes positioned at various locations of the scalp, capturing brain activities with millisecond-level temporal resolution. Therefore, EEG signals are represented in the form of high-dimensional, sequential data. However, this representation limits a direct insights into the patterns of different sleep stages. To address this issue, one of the most common and effective approaches is to employ dimensionality reduction algorithms to map the high-dimensional EEG data into a lower-dimensional space for visualization. This mapping incurs a certain degree of information loss. Hence, it necessitates that the dimensionality reduction algorithms must be capable of sufficiently preserving relevant information.

Machine learning algorithms, especially deep learning, serves as powerful mapping functions, demonstrating exceptional performance in transforming high-dimensional data into lower-dimensional representations. While these methods have already offered unprecedented accuracy and predictive capabilities, the comprehension of the mathematical supports behind them is still not thorough. This shortfall is of minimal concern from an engineering perspective, where the primary objective is to achieve satisfactory results. However, their contributions to fundamental research, particularly to physiology, have not been significant. Without a comprehensive understanding of the intermediate processes, it is almost impossible to conduct any rigorously logical judgement from the experimental results. Several explainable models whose intermediate processes are transparent contribute to fundamental research to a certain extent. However, these models are limited by numerous external factors, such as noises in the input data and the requirements on the substantial computational resources. Moreover, the performances of these explainable models depend extremely on the quality of input features, further limiting their usage in elucidating the complex dynamics of EEG signal processing. Similar to the situation with machine learning in the EEG signal processing field, numerous novel representations of sequential data have been imported, such as graphs. The features extracted from these data structures frequently lack a thorough examination of their effectiveness before being utilized as input data. This gap is evident as such a practice may not only undermine the potential performance of the following algorithms but also constrain the analysis of the output results.

Topological Data Analysis (TDA) is an advanced data analysis method based on topology and geometry, whose core idea is to understand the essential characteristics and inherent structures by identifying the relationships within data. Two most notable TDA algorithms are the Mapper algorithm~\cite{singh2007topological} and the persistent homology~\cite{edelsbrunner2008persistent}, which have been proved to be successful in various fields such as biology, physics, and medicine~\cite{carlsson2021topological}. The Mapper algorithm maps high-dimensional data space to a lower-dimensional representation, utilizing a filter function, and constructs a simplified topological graph to reveal the fundamental structure and patterns of the data. In comparison, persistent homology identifies and quantifies the “persistence” of topological features by analyzing the emergence and disappearance of voids across different scales, thereby revealing the intrinsic structure of the data. Traditional processing methods for time series data are either frequency or time-domain analysis approaches. However, there are abundant structural characteristics existing in EEG signals unexplored, which can be supplemented by TDA.

The objective of our research is to leverage explainable models for transforming high-dimensional sequential EEG signals into visualizable low-dimensional representation, facilitating the use of EEG signals in foundational research of exploring sleep patterns. The performance of explainable models is generally limited by input features and is highly sensitive to noise and outliers. To enhance the robustness of explainable models, we introduce the Recursive Feature Elimination with Cross-Validation (RFECV) feature selection algorithm to perform an initial selection of input features, aiming to reduce ineffective and redundant information. Then we integrate this algorithm with dimensionality reduction techniques to form the final two-stage hierarchical and explainable feature selection framework. To enrich the variety of input features and thereby enhance model performance, we apply TDA to extract structural features from EEG signals. Traditional spectral-temporal features focus more on the variations of signal values themselves, neglecting the global structural information of the data. During the experiment part, we conduct persistent homology visual analyses for different sleep stages. The differences observed indicated that TDA features carry informative and effective information for sleep classification. Subsequently, TDA features and traditional spectral-temporal features serve as input data for the two-stage hierarchical and explainable feature selection framework. The classification results demonstrate that TDA features cannot replace traditional spectral-temporal features completely, but can serve as an effective complement to them. Additionally, we utilize the proposed two-stage framework to reduce the high-dimensional EEG signals to two dimensions for visualization purpose, and effective clustering of the five different sleep stages. In conclusion, the import of TDA features and the proposed two-stage hierarchical framework have demonstrated excellent performances in dimensionality reduction of EEG signals and classification of sleep stages. The use of explainable models has made the intermediate results clear, which can be utilized for foundational research in exploring sleep patterns.

The subsequent sections of the paper are organized as follows: Section II introduces the related works on the EEG signal features and dimensionality reduction algorithms, along with the core concepts of TDA. Section III presents the datasets used, the proposed algorithm framework, and the different input features. Section IV provides a comprehensive analysis of the algorithm framework in terms of classification performance and visualization in two-dimensional space, as well as a visual analysis of the persistent homology for different sleep stages. Finally, Section V summarizes the essential content of the paper and discusses the future work.

\section{Related Works}

\subsection{EEG Signal Features}
EEG data is a type of high-dimensional time-series data with complex spatial distribution, containing rich sleep-related information across multiple time scales. Therefore, various methods for extracting features from high-dimensional EEG signals have been proposed~\cite{subasi2007eeg, fathima2011wavelet, geng2011eeg}. The extracted EEG signal features can generally be divided into three types: time-domain, frequency-domain, and time-frequency domain features. Time-domain features, as one of the most fundamental and effective types of characteristics, are widely used due to their ease of operation and information richness~\cite{aboalayon2014multi, chinara2021automatic}.  Frequency-domain features are extracted by passing EEG signals through filters to separate them into different frequencies~\cite{stancin2021eeg, huang2022classification, chen2022fusing}. These signals of various frequencies can be used to study frequency-sensitive physiological and pathophysiological phenomena, such as sleep~\cite{you2022automatic} and Alzheimer's disease~\cite{li2021feature}. Time-frequency domain features are also information-rich, incorporating both the time-domain and frequency-domain characteristics, but require significant computational resources and cannot fully replace time-domain and frequency-domain features~\cite{bhuvaneshwari2021classification, hag2021eeg, chen2023eeg}.

\begin{figure}[htb] 
	\centering
	\includegraphics[width=\linewidth]{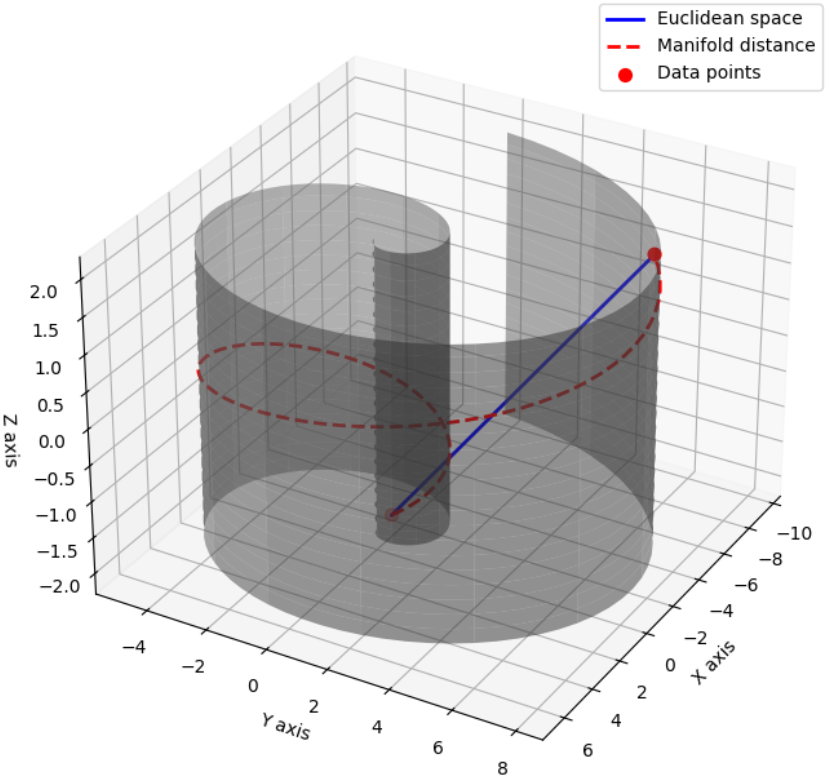}
	\caption{Manifold distance and Euclidean space.}
	\label{fig:manifold}
\end{figure}

\subsection{Dimensionality Reduction Algorithms}
With significant advancements in hardware, the performance of deep neural networks has been greatly enhanced. In the field of sleep stage detection using EEG signals, the introduction of DeepsleepNet~\cite{supratak2017deepsleepnet} has led to the adoption of increasingly advanced deep learning models, such as transformers~\cite{shi2021transformer} and graph neural networks~\cite{cai2020graph}. These algorithms demonstrate exceptional performance in terms of classification accuracy and generalization ability. However, the opaque intermediate processes in the deep models limit their usage for whitebox analysis in fundamental research. Other notable works utilized explainable models with EEG signals to make significant contributions to sleep stage detection, such as PCA~\cite{gilmour2010manual}, ICA~\cite{huang2022key}, and Isomap~\cite{krivov2016dimensionality}. Nevertheless, the performance of these algorithms is inferior to those achieved by deep learning. To balance model interpretability and the effectiveness of sleep classification, we propose an effective, yet interpretable, model.

\subsection{Topological Data Analysis}
TDA is a collection of advanced algorithms based on topology, aiming to discover the hidden structures inherent in data itself. Over the past decade, TDA has found extensive applications in many sectors, particularly in finance. With methods such as graph theory, which explore the structures of data itself, gaining success in the EEG signal processing, many efforts have begun to integrate TDA algorithms into EEG signal analysis~\cite{manjunath2023topological, xu2021topological, yamanashi2021topological}. These achievements provide valuable insights for our research. Note that there are currently no studies that comprehensively discuss how to use TDA algorithms for sleep classification based on EEG signals. Our work aims to fill in this gap. The rest of this subsection gives a brief introduction to persistent homology, which focuses exclusively on the core concepts and ideas in this area. For more comprehensive reviews of the underlying mathematical principles, readers are encouraged to consult references~\cite{mendelson1990introduction, nash1988topology, ghrist2018homological, maclane2012homology} for group theory and~\cite{rabadan2019topological, ghrist2014elementary, boissonnat2018geometric, chazal2021introduction} for insights into TDA.

\subsubsection{Manifold}
The concept of manifold refers to a topological space locally resembling Euclidean space. In particular, an $n$-dimensional manifold is defined as a topological space where every point contains a neighborhood being homeomorphic to an open subset of $n$-dimensional Euclidean space. TDA is based on the assumption that data samples are derived from sampling an unknown high-dimensional manifold, and it is the aim of TDA to extract topological information from the underlying manifold. One of the biggest differences between manifolds and ordinary geometric sets lies in the understanding of the distance between two points in space. As shown in Fig. 1, two close points in terms of Euclidean distance may necessitate circling twice along the curve for distance assessment within the manifold distance. In our work, EEG signals are sequential data which can be represented in Euclidean space, thus making TDA applicable for the analysis.

\begin{figure}[h] 
	\centering
	\includegraphics[width=\linewidth]{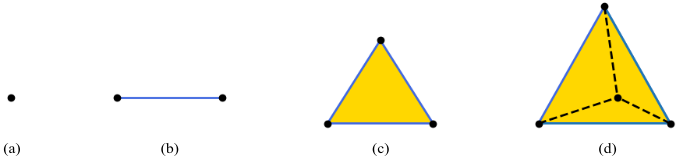}
	\caption{Examples of zero-dimensional (a), one-dimensional (b), two-dimensional (c), and three-dimensional simplices (d).}
	\label{fig:simplices}
\end{figure}

\begin{figure}[h]
	\centering
	\includegraphics[width=\linewidth, height=0.8\textheight, keepaspectratio]{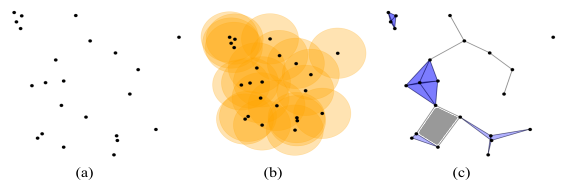}
	\caption{Example of a VR complex. (a) Data points in $X$. (b) Each point attached with a ball of radius $\epsilon$. (c) Clusters in the VR complex $V(X, \epsilon)$.}
	\label{fig:VRC}
\end{figure}

\subsubsection{Simplicial Complex}
\begin{figure*}[!h]  
	\centering           
	\includegraphics[width=\textwidth]{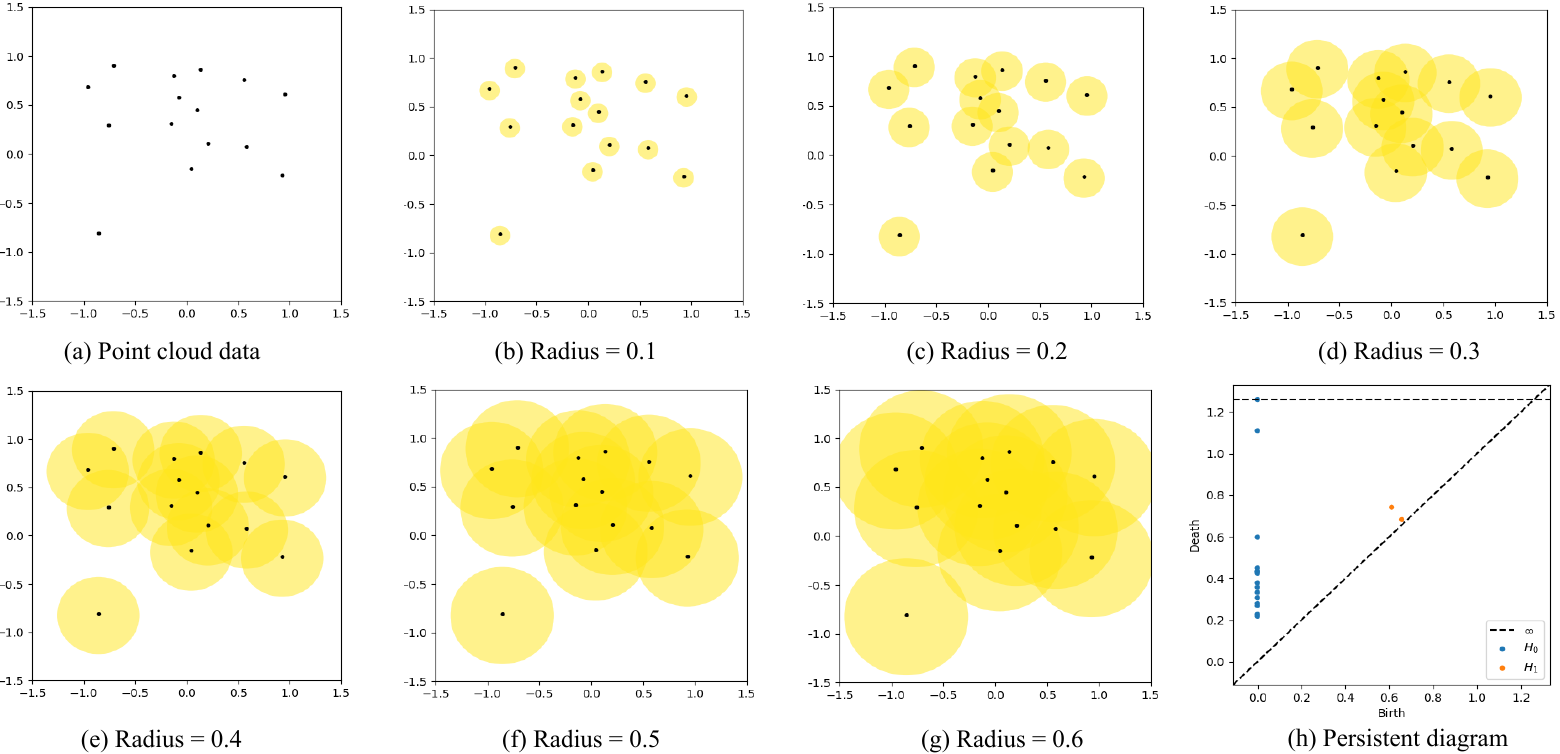}  
	\caption{The process of Rips complex filtration. Subfigures (a)-(g) are schematic diagrams at different radii of the data points, where overlapping regions will have a deeper color. Subfigure (f) is the persistent diagram of the data points in subfigure (a) as their radius increases from $0$ to $\infty$.}  
	\label{fig:routine}  
\end{figure*}

A simplicial complex is a type of high-dimensional geometric structure analogous to the graph, employed to quantify the shape of the data. In persistent homology, observed data points serve as the vertices of the simplicial complexes. The way of encoding connections between vertices is dependent on the definition of distance in measured space. For easy understanding, only the most popular one, the Vietoris-Rips (VR) complex~\cite{carlsson2006algebraic}, is discussed within this paper.

The VR complex is constructed by data points existing in the metric space. A metric $\rho$ on a set $X$ is a function mapping any two elements from $X$ into the set of real numbers $\mathbb{R}$, which must obey the following properties: 

(i) For any $x, y \in X$, $\rho(x, y) \geq 0$, and $\rho(x, y) = 0$ iff $x = y$; 

(ii) For any $x, y \in X$,  $\rho(x, y) = \rho(y, x)$;

(iii) For any $x, y, z \in X$, $\rho(x, z) \leq \rho(x, y) + \rho(y, z)$. 

A metric space is defined as $(X, \rho)$, where $\rho$ is the metric on $X$. As illustrated in Fig. 3, a VR complex is denoted as $V(X, \epsilon)$, where $\epsilon > 0$ is the measure determining the scale of the VR complex~\cite{chambers2010vietoris}. Every VR complex can be decomposed into several basic blocks named simplices. A $k$ dimensional simplex, or $k$-simplex, is defined as the convex hull of $k+1$ affinely independent points satisfying that the distance between any two points of these $k+1$ points is no greater than $\epsilon$. Examples of the zero-dimensional, one-dimensional, two-dimensional, and three-dimensional simplices in Euclidean space are shown in Fig. 2. For instance, a two-dimensional simplex forms a triangular area, while a three-dimensional simplex constitutes a tetrahedral volume. Hence, a VR complex, $V(X, \epsilon)$, can be filtrated into a nested sequence of sub-complexes, $\emptyset = K^0 \subset K^1 \subset … \subset K^{m+1} = V(X, \epsilon)$, where $K^{i+1} = K^i \cup \Delta^i$ where $\emptyset$ represents the empty set, $\Delta^i$ is the set of $i$-dimensional simplices of $X$, $i = 0, 1, ..., m$, and $m$ is the dimension of the metric space [15]. With $\epsilon$ increasing from $0$ to $\infty$, the range of $V(X,\epsilon)$ may change for different values of $\epsilon$. This results in a sequence of VR complexes for increasing values of $\epsilon$, and such a process of obtaining the sequence of VR complexes is called the VR filtration.

\subsubsection{Persistent Homology}
The core idea of persistent homology is to analyze the underlying manifold shape of data by calculating the voids formed during the construction process of simplicial complexes. A void is an uncovered region enclosed by simplices, which serve as boundaries of the void. As demonstrated in Fig. 2, with the distance of the ball $\epsilon$ increasing, connections gradually emerge between points and some enclosed areas start to appear within the simplicial complex. Differentiated from the enclose region belonging to the simplices (the purple region), a void refers to the enclosed gray region in the rightmost subfigure of Fig. 2.

Persistent homology analyzes the entire filtration of simplicial complexes from the intial stage where data points are isolated to the final stage where any two data points are connected. During this procedure, the birth time, death time, and the numbers of voids of different dimensions are regarded as essential topological information. In particular, the numbers of the voids are defined as the Betti numbers, such that the number of voids in a certain $k$ dimension is denoted as $H_k$. A pair of birth time $\epsilon_{\min}$ and death time $\epsilon_{\max}$ corresponds to the epsilon values at which the void starts to appear and disappear (covered). The range $[\epsilon_{\min}, \epsilon_{\max}]$ is named as the persistence intervals, each of which contributes to the Betti numbers. The collection of all persistence intervals serve as descriptive information for the underlying manifold. This provides insights into the dimension in which voids are present within the data and the range of values over which voids persist.

\subsubsection{Persistent Diagrams}
The persistence intervals are essentially pairs of real numbers not easily analyzed. There are two intuitive methods for visualizing persistence intervals: persistent diagrams and barcodes. They are basically equivalent but different representations of intervals. Hence, only the more popular one, persistent diagram, is introduced here. As displayed in the subfigure (h) of Fig. 4, the $x$-axis and $y$-axis represent the birth and death times of voids respectively, thus representing each interval as the point $(\epsilon_{\min}, \epsilon_{\max})$. As the death time of any void cannot be earlier than its birth time, no points exist below the dashed line $y=x$. Besides, the vertical distance of a point from the line $y=x$ is directly proportional to the length of the interval. This means that the further a point is from the line $y=x$, the more persistent the features carried by that point are.

\section{Methods}
In this section, we introduce the datasets used in the experiments and explain the techniques for extracting two types of features: spectral-temporal features and persistent homology features. Then, we further elaborate on the proposed Two-Stage Hierarchical and Explainable Feature Selection Framework for dimensionality reduction, which can be divided into two parts: the feature selection part and the dimensionality reduction algorithms.

\subsection{Dataset}
\textbf{Sleep-EDF}~\cite{physionet-sleepedfx}: Each raw EEG signal was collected from the Caucasian aged from 21-35 years, sampled at the frequency of 100 Hz in 30 sec intervals. Among the participants, some experienced mild sleep disturbances, leading to fewer or even no data labels during the sleep stage N3.

\subsection{Feature Extraction}

We classify the features extracted from the EEG signals into two types: spectro-temporal features and structure-based features. The features obtained from spectral-temporal analysis can be divided into time-domain features, frequency-domain features, and time-frequency domain features. Following the summary in~\cite{chung2021persistent, jenke2014feature}, 33 features were extracted as spectro-temporal features. For structure-based analysis, we employed persistence statistics~\cite{csen2014comparative} to analyze the extracted persistent intervals, thereby obtaining 18 features. Hence, a total of 51 feature candidates are used as input data into the feature selection algorithm.

\subsubsection{Spectral-Temporal Features}

 As shown in Table I, we use $X$ to represent a complete and continuous segment of EEG signal data and $\{x_1, x_2, …, x_{n}\} \subseteq X$ to represent a sequence of $n$ data points in $X$. Thirty-three features across three different domains are calculated using fifteen different algorithms. Spectral-temporal features combine frequency (spectral) information and temporal information of signals to comprehensively describe the signals's dynamic characteristics, which are of significant importance for the classification of sleep stages~\cite{fathima2011wavelet, geng2011eeg}. Sleep is a dynamic process: an individual transitions multiple times between sleep stages throughout the sleep. Different sleep stages have distinct spectral signatures. For example, delta waves are the predominant brain electrical activities during the N3 stage. On the other hand, the cyclical variations of sleep can be precisely captured through time-domain analysis. Besides, time-frequency domain features serve as an effective supplement to the aforementioned feature types.

\subsubsection{Persistent Homology Features}

Traditional spectral-temporal analysis is susceptible to noise interference and typically relies on linear or quasi-linear assumptions. Therefore, this approach fails to capture nonlinear characteristics and neglects the structural information within the data. Persistent homology features, which are based on analyzing the topological structure of the data, address the forementioned problems. In this work, we consider persistence statistics (PS) to serve as critical information to classify sleep stages in addition to spectral-temporal features.
We define the persistent interval as $T = \{[\epsilon_{\min}^i, \epsilon_{\max}^i], i = 1, 2, ...\}$. We denote the midlife persistence as $T_\mathrm{m} = \{(\epsilon_{\min} + \epsilon_{\max})/2\ |\ [\epsilon_{\min}, \epsilon_{\max}] \in T\}$ and the lifespan persistence is denotes as $T_\mathrm{l} = \{\epsilon_{\max} - \epsilon_{\min}\ |\ [\epsilon_{\min}, \epsilon_{\max}] \in T \}$. After that, eight classical statistics are calculated, including mean, standard, deviation, skewness, kurtosis, 25th percentile, 50th percentile, 75th percentile, and entropy of $T_\mathrm{m}$ and $T_\mathrm{l}$, specified as mean($T_\mathrm{m}$), std($T_\mathrm{m}$), sk($T_\mathrm{m}$), ku($T_\mathrm{m}$), $P_{25}(T_\mathrm{m})$, $P_{50}(T_\mathrm{m})$, $P_{75}(T_\mathrm{m})$, $H(T_\mathrm{m})$, mean($T_\mathrm{l}$), std($T_\mathrm{l}$), sk($T_\mathrm{l}$), ku($T_\mathrm{l}$), $P_{25}(T_\mathrm{l})$, $P_{50}(T_\mathrm{l})$, $P_{75}(T_\mathrm{l})$, and $H(T_\mathrm{l})$, respectively.

\begin{table*}[p]
	\centering
	\caption{List of all the features considered in this work}
	\renewcommand{\arraystretch}{1.5} 
	\footnotesize 
	\begin{tabularx}{\textwidth}{|c|>{\centering\arraybackslash}X|c|c|} 
		\hline
		\textbf{Feature} & \textbf{Feature detail and description} & \textbf{Domain} & \textbf{Dimension} \\ \hline
		Number of zero crossings ($N_\mathrm{zc}$) & 
		\begin{tabular}[c]{@{}c@{}} 
			A zero crossing ocurs in one of the following conditions: \\
			$(x_{n-1}<0 \text{ and } x_n>0) \text{ or } (x_{n-1}>0 \text{ and } x_n<0) $ \\ $\text{ or } (x_{n-1} \neq 0 \text{ and } x_n=0)$
		\end{tabular} & Time & 1 \\ \hline
		
		\begin{tabular}[c]{@{}c@{}}Hjorth parameters: \\ mobility ($M_\mathrm{H}$),\\ complexity ($C_\mathrm{H}$),\\ and activity ($A_\mathrm{H}$)\end{tabular} & 
		\begin{tabular}[c]{@{}c@{}} 
			$A_\mathrm{H} = \sigma_0^2,$ $M_\mathrm{H} = \frac{\sigma_1}{\sigma_0},$ \\
			$C_\mathrm{H} = \sqrt{\frac{\sigma_2\sigma_1}{\sigma_0^2}}$ \\ 
			where $\sigma_0$, $\sigma_1$, and $\sigma_2$ represent the standard deviations of the 0th, 1st, and 2nd \\ derivatives of $X$
		\end{tabular} & Time & 3 \\ \hline
		
		\begin{tabular}[c]{@{}c@{}}Traditional statistical features:\\ minimum value ($x_\mathrm{min}$),\\ maximum value ($x_\mathrm{max}$),\\ arithmetic mean ($\overline{x}$),\\ standard deviation ($\sigma$),\\ variance ($\sigma^2$),\\ skewness ($\gamma_1$),\\ kurthosis ($\beta_2$),\\ and median ($P_{50}$)\end{tabular} &
		\begin{tabular}[c]{@{}c@{}} 
			$x_\mathrm{min}=\min\{x_i\ |\ i=1, 2, ..., n\}$,
			$x_\mathrm{max}=\max\{x_i\ |\ i=1, 2, ..., n\}$,\\ 
			$\overline{x}=\frac{1}{n} \sum_{i=1}^n x_i$,
			$\sigma = \sqrt{\frac{1}{n} \sum_{i=1}^n (x_i - \overline{x})^2}$,\\ 
			$\gamma_1 = \frac{1}{n} \sum_{i=1}^{n} \left( \frac{x_i - \overline{x}}{\sigma} \right)^3
			$, 
			$\beta_2 = \frac{1}{n} \sum_{i=1}^{n} \left( \frac{x_i - \overline{x}}{\sigma} \right)^4$, \\
			$P_{50} = 
			\begin{cases} 
				x_{\left(n+1\right)/2}, & \text{if } n \text{ is odd} \\
				(x_{\left(n/2\right)} + x_{\left(n/2+1\right)})/2, & \text{if } n \text{ is even}
			\end{cases}
			$
		\end{tabular} & Time & 8 \\ \hline
		
		Petrosian fractal dimension ($D_\mathrm{P}$) &
		\begin{tabular}[c]{@{}c@{}}
			$
			D_\mathrm{P} = \frac{\log_{10} N}{\log_{10} N + \log_{10} \left(\frac{N}{N + 0.4N_{\delta}}\right)}
			$ \\
			where $N$ represents the total number of data points in the sequence, $N_{\delta}$\\ represents the number of times the absolute difference between consecutive data\\ points satisfies $|x(n) - x(n-1)| \geq \delta$, and $\delta$ is a predefined threshold
		\end{tabular} & Time & 1 \\ \hline
		
		Mean teager energy ($\overline{E}_\mathrm{T}$) &
		\begin{tabular}[c]{@{}c@{}}
			$
			\overline{E}_\mathrm{T} = \frac{1}{N} \sum_{n=1}^N \left( x_n^2 - x_{n+1}x_{n-1} \right)
			$
		\end{tabular} & Time & 1 \\ \hline
		
		Mean energy ($\overline{E}$) &
		\begin{tabular}[c]{@{}c@{}}
			$
			\overline{E} = \frac{1}{N} \sum_{n=1}^N x_n^2
			$
		\end{tabular} & Time & 1 \\ \hline
		
		Mean curve length ($\overline{L}_\mathrm{c}$) &
		\begin{tabular}[c]{@{}c@{}}
			$\overline{L}_\mathrm{c} = \frac{1}{N-1} \sum_{n=1}^{N-1} \sqrt{1 + (x_{n+1} - x_{n})^2}$
		\end{tabular} & Time & 1 \\ \hline
		
		Hurst exponent ($H_\mathrm{e}$) &
		\begin{tabular}[c]{@{}c@{}}
			$
			H_\mathrm{e} = \lim\limits_{T \to \infty} \frac{\log(R/S)}{\log(T)}
			$\\ where $T$ represents the length of the considered time series, $R$ represents the\\ maximum range of the time series for length $T$, and $S$ represents the standard\\ deviation for the corresponding range
		\end{tabular} & Time & 1 \\ \hline
		
		Band power ($P_\mathrm{band}$) &
		\begin{tabular}[c]{@{}c@{}}
			$P_\mathrm{band} = \frac{1}{N}\sum_{i=1}^N |x^b_i|^2
			$;\\ we consider four types of waveband: delta (1-4 Hz), theta (4-8 Hz), alpha\\ (8-13 Hz), and beta (13-30 Hz) for all frequency-domain features; \\ $\left| x^b_i \right|$ represents the amplitude of the band pass filtered signal
		\end{tabular} & Frequency & 4 \\ \hline	
		
		Power spectral density ($S(f_k)$) &
		\begin{tabular}[c]{@{}c@{}}
			$S(f_k) = \frac{1}{N} \sum_{k}|X_{f_k}|^2
			$\\ where $k$ represents the frequency index and $|X_{f_k}|$ represents the\\ frequency-domain representation of the time series $\{x_i\ |\ i=1, 2, ...\}$ at the\\ corresponding frequency $f_k$ after the Fourier transform
		\end{tabular} & Frequency & 4 \\ \hline	
		
		Relative power ($P_\mathrm{rel}$) &
		\begin{tabular}[c]{@{}c@{}}
			$P_\mathrm{rel} = \frac{\sum_{k_1} S(f_{k_1})}{\sum_{k} S(f_k)}$\\
			where $k_1$ represents the index of the target waveband and $k$ represents the index\\ of all wavebands
		\end{tabular} & Frequency & 4 \\ \hline	
		
		Peak frequency ($f_\mathrm{peak}$) &
		\begin{tabular}[c]{@{}c@{}}
			$
			f_\mathrm{peak} = \Delta f \cdot \arg\max S(f_k)
			$\\ where $\Delta f$ represents the frequency resolution
		\end{tabular} & Frequency & 4 \\ \hline	
		
		Spectral entropy ($H_\mathrm{s}$) &
		\begin{tabular}[c]{@{}c@{}}
			$H_\mathrm{s} = -\sum_{k} P(f_k) \log_2 P(f_k)$\\
			where $P(f_k)$ represents the result of normalizing $S(f_k)$
		\end{tabular} & Frequency & 4 \\ \hline	
		
		Discrete wavelet transform  &
		\begin{tabular}[c]{@{}c@{}}
			$
			X_{j, k} = \sum_{i=1}^{N} x_i\psi_{j,k}[i]
			$,\\ where $X_{j, k}$ represents the wavelet coefficients obtained through the discrete\\ wavelet transform and $\psi_{j,k}[i]$ is the wavelet function at scale $j$ and\\ translation $k$
		\end{tabular} & Time-frequency & 16 \\ \hline	
		
		Persistence statistics &
		\begin{tabular}[c]{@{}c@{}}
			mean($T_\mathrm{m}$), std($T_\mathrm{m}$), sk($T_\mathrm{m}$), ku($T_\mathrm{m}$), $P_{25}(T_\mathrm{m})$, $P_{50}(T_\mathrm{m})$, $P_{75}(T_\mathrm{m})$, $H(T_\mathrm{m})$,\\ mean($T_\mathrm{l}$), std($T_\mathrm{l}$), sk($T_\mathrm{l}$), ku($T_\mathrm{l}$), $P_{25}(T_\mathrm{l})$, $P_{50}(T_\mathrm{l})$, $P_{75}(T_\mathrm{l})$, $H(T_\mathrm{l})$;\\
			calculation formulas are similar to traditional statistical features
		\end{tabular} & Topology & 16 \\ \hline	
		
	\end{tabularx}
	
\end{table*}

\subsection{Two-stage Hierarchical Feature Selection Framework for Dimensionality Reduction}

The two-stage hierarchical framework can be divided into two parts: feature selection algorithm and dimensionality reduction algorithm. The entire algorithm process is shown in Fig. 5. In our work, the Recursive Feature Elimination algorithm was used for feature selection. We compare three algorithms: PCA, t-SNE, and UMAP, in terms of numerical results and visualization outcomes on EEG data dimensionality reduction.

\begin{figure*}[!htb]  
	\centering           
	\includegraphics[width=\textwidth]{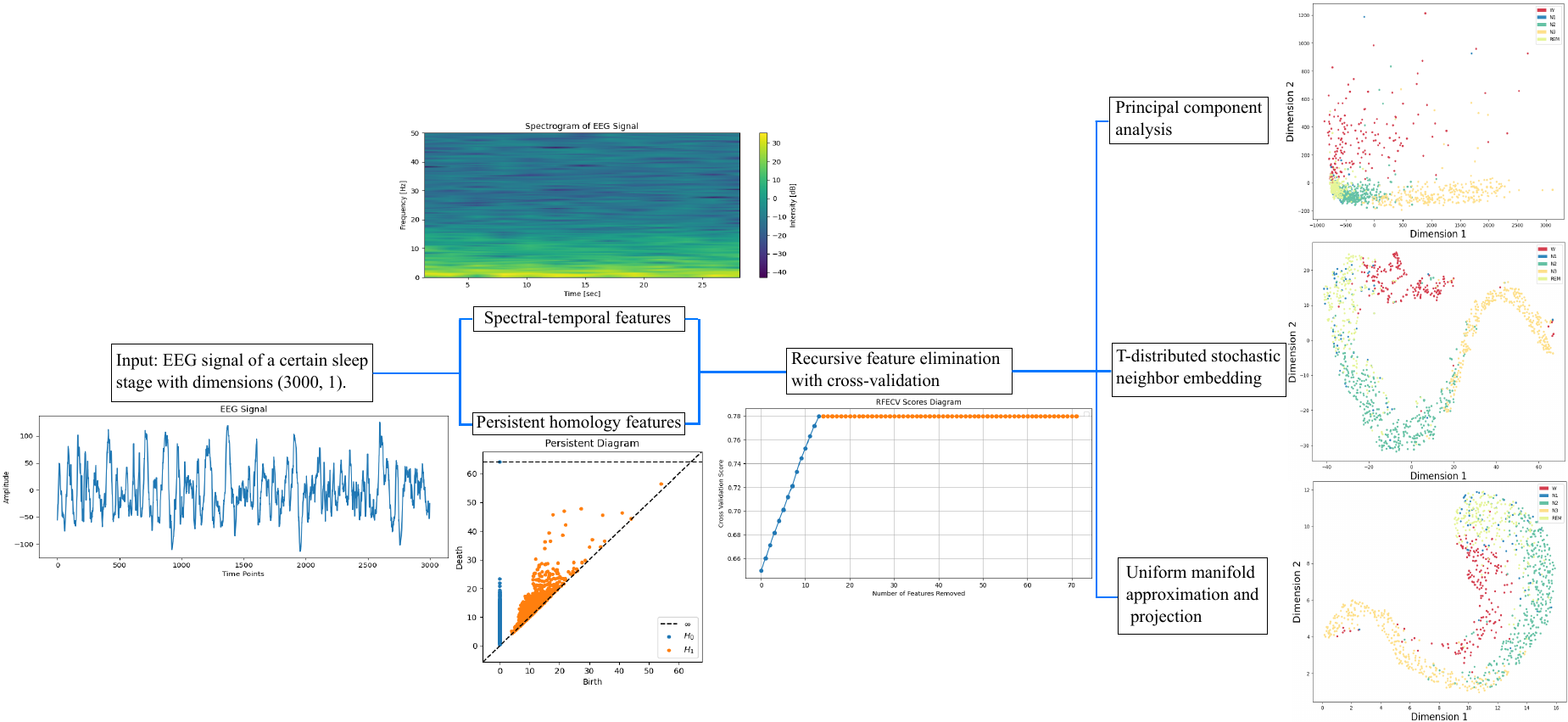}  
	\caption{Two-stage hierarchical feature selection framework for dimensionality reduction.}  
	\label{fig:routine}  
\end{figure*}

\subsubsection{Recursive Feature Elimination With Cross-Validation}

Recursive Feature Elimination with Cross-Validation (RFECV) is a wrapper feature selection algorithm, which means that an independent machine learning algorithm should serve as the core of the method, wrapped by RFECV. In our work, K-Nearest Neighbors algorithm (KNN) was used as for the central part of RFECV. In particular, the accuracy of various features based on the KNN algorithm were used as the metric to determine whether to retain a certain feature. As illustrated in Algorithm I, we combined RFECV with the greedy algorithm that iteratively removes features to maximize the metric improvement. When converged, the resultant feature set may include some redundancy. However, this is not problematic as the subsequent stage involves the use of a dimensionality reduction algorithm to eliminate redundant information. By conducting a preliminary selection with RFECV, features with weak correlations are removed, and this reduces the workload of the dimensionality reduction algorithm in the second stage.

\vspace{5mm}
\noindent 
\begin{tabular}{|L{8.5cm}|} 
	\hline
	\textbf{Algorithm 1:} Recursive feature elimination with cross-validation (RFECV) \\
	\hline
	\textbf{Input:} Training data feature candidates $x$, training data labels $y$, number of neighbors for the KNN classifier $n\_neighbors$. \\
	\textbf{Process:} \\ Initialize feature set $F$ to all features. \\ Initialize $current\_accuracy$ to $0$. \\ Initialize feature set $best\_features$ to empty set. \\
	Initialize $improvement\_found$ to False. \\
	\textbf{For} each feature $f$ \textbf{in} $F$: \\
	\quad Create a feature subset $temp\_F$ containing all \\ \quad features in $F$ except $f$. \\ \quad Perform cross-validation using $temp\_F$ and  KNN \\ \quad classifier with $n\_neighbors$. \\ \quad Calculate the cross-validation score $new\_accuracy$.
	\\ \quad \textbf{If} $new\_accuracy > current\_accuracy$: 
	\\ \quad \quad Update $current\_accuracy$ to $new\_accuracy$. 
	\\ \quad \quad Set $best\_features$ to $temp\_F$. 
	\\ \quad \quad Set $improvement\_found$ to True.
	\\ \quad \textbf{If} $improvement\_found$ \textbf{is} False: 
	\\ \quad \quad \textbf{Break} the loop. 
	\\ \quad Update $F$ to $best\_features$. \\
	\textbf{Output:} Feature set $best\_features$. \\
	\hline
\end{tabular}
\vspace{5mm}

\subsubsection{Principal Component Analysis}

Principal Component Analysis (PCA) is a linear dimensionality reduction technique which captures the directions of maximum variance in data by transforming the data into a new coordinate system. It extracts the most information-rich features from a large dataset while preserving the most relevant information from the initial dataset. The core idea of PCA is to represent the information content of the data into a small number of uncorrelated variables, called principal components. These principal components are linear combinations of the original feature variables, containing the maximum data variance compared with other possible combinations. The PCA is implemented as follows: (i) standardize the data; (ii) compute the covariance matric of the data; (iii) calculate the eigenvectors and eigenvalues of the covariance matrix; (iv) sort the eigenvalues and form a feature vector from the eigenvectors; and (v) reconstruct the data utilizing the eigenvectors. 

The linear and interpretable nature of the PCA enables it to serve as an effective and convenient tool for denoising, dimensionality reduction, and visualization in various research areas, including EEG signal processing. The absence of any hyperparameters in the algorithm also greatly conserves computational resources. In this paper, PCA was used as a benchmark for comparison with other algorithms, and with it, we fully evaluated the performance of the manifold-based dimensionality reduction algorithms.


\subsubsection{t-Distributed Stochastic Neighbor Embedding}

t-Distributed Stochastic Neighbor Embedding (t-SNE) is a nonlinear dimensionality reduction algorithm that visualizes high-dimensional data by embedding the data onto a lower-dimensional manifold. Originally, based on the stochastic neighbor embedding proposed by Hinton and Roweis~\cite{hinton2002stochastic}, the t-distributed variant was developed by van der Matten~\cite{van2008visualizing}. The core of the algorithm is to map high-dimensional data into a low-dimensional space while preserving the relative distances in high-dimensional space. It first calculates the similarities between data points in high-dimensional space and reconstructs these similarities in the low-dimensional space. During this process, the manifold assumption is used to guide and optimize the layout of data points in the low-dimensional space.

As illustrated in the Algorithm 2, The t-SNE algorithm includes two main stages. In the first stage, t-SNE constructs a probability distribution over pairs of high-dimensional objects, introducing similarity as probability. For each point $x_i$, the conditional probability $p_{j|i}$ is calculated with respect to another point $x_j$. This probability represents the likelihood that $x_j$ is chosen as a neighbor of $x_i$. This is typically modeled using a Gaussian distribution: 
\[p_{j|i} = \frac{e^{-\|x_i - x_j\|^2 / (2\sigma_i^2)}}{\sum_{k \neq i} e^{-\|x_k - x_i\|^2 / (2\sigma_i^2)}}\] where $\|\cdot\|$ represents the Euclidean norm. To ensure that the model is sensitive to both small and large distances in high-dimensional data, $p_{ij}$ is typically set as the symmetrized version of $p_{j|i}$ and $p_{i|j}$:
\[p_{ij} = \frac{p_{j|i} + p_{i|j}}{2N}.\]
In the second stage, t-SNE calculates, in the low-dimensional space, a probability distribution $q_{ij}$ to describe the similarity between points $y_i$ and $y_j$ (which are low dimensional representations of $y_i$ and $y_j$):
\[q_{ij} = \frac{(1 + \|y_i - y_j\|^2)^{-1}}{\sum_{k, l,\ k \neq l} (1 + \|y_k - y_l\|^2)^{-1}}.\]
The goal of t-SNE is to make the probability distribution $q_{ij}$ in the low-dimensional space as close as possible to the probability distribution $p_{ij}$ in the high-dimensional space. This is achieved by minimizing the Kullback-Leibler (KL) divergence between the two distributions:
\[D_\mathrm{KL} = \sum_{i, j,\ i \neq j} p_{ij} \log \frac{p_{ij}}{q_{ij}}.\]
The optimization of this loss function typically uses the gradient descent method.

\vspace{5mm}
\noindent 
\begin{tabular}{|L{8.5cm}|} 
	\hline
	\textbf{Algorithm 2:} t-distributed stochastic neighbor embedding (t-SNE) \\
	\hline
	\textbf{Input:} High-dimensional data $X$, dimensionality of the reduced space $n\_components$, perplexity value $perlexity$, number of optimization epochs $n\_iter$, learning rate $learning\_rate$.
	\\
	\textbf{Process:} 
	\\ Compute pairwise distances matrix $D$ for high-dimensional data $X$.
	\\ Use binary search to find $\sigma_i$ for each point $i$ that approximates the given perplexity. 
	\\ Calculate conditional probabilities $p_{j|i}$ in the high-dimensional space.
	\\ Symmetrize $p_{ij}$ to make it undirected.
	\\ Initialize low-dimensional embeddings $Y$ to random values.
	\\ \textbf{For} each iteration $t = 1$ \textbf{in} $n\_iter$:
	\\	\quad Compute pairwise distances matrix $D_Y$ in the \\ \quad low-dimensional space.
	\\	\quad Calculate probabilities $q_{ij}$ using the t-distribution.
	\\	\quad Compute the gradient of the KL divergence.
	\\	\quad Update low-dimensional embeddings.
	\\ \textbf{Output:} The low-dimensional embeddings $Y$.\\
	\hline
\end{tabular}
\vspace{5mm}

\subsubsection{Uniform Manifold Approximation and Projection}

UMAP is a nonlinear dimensionality reduction technique that reduces and visualizes high-dimensional data by preserving the underlying manifold structure within the data~\cite{mcinnes2018umap}. Through manifold learning, both local and global structures of the data can be preserved. As shown in Algorithm 3, the algorithm can be divided into four steps. Firstly, a neighborhood graph of the data is constructed based on the determined hyperparameter $n\_neighbors$. For each point, the closest $n\_neighbors$ points are selected as neighbors to form a directed graph, where each point points to its $n\_neighbors$ nearest neighbors. Secondly, the original distance metric is transformed into connection strengths using a Gaussian kernel. For each point $i$, an appropriate local scale $\sigma_i$ is found such that the distribution of connection strengths between point $i$ and its $n\_neighbors$ neighbors approximates a Gaussian distribution. Subsequently, the weight $w_{ij}$ from point $i$ to point $j$ is calculated using a Gaussian function: \[e^{-\frac{d_{ij}^2}{2\sigma_i^2}}\]
where $d_{ij}$ is the distance between points $i$ and $j$. Thirdly, the connection strengths in the neighborhood graph are used to find the optimal position for each point in a lower-dimensional space. Finally, the goal of UMAP is to find a low-dimensional representation that preserves the structure of the high-dimensional space as well as in the lower-dimensional space. This involves minimizing the following cost function: 
\[
	L = \sum_{i,j} \left(w_{ij} \log \frac{w_{ij}}{w'_{ij}} + (1 - w_{ij}) \log \frac{1 - w_{ij}}{1 - w'_{ij}}\right).
\]
In this context, $w'_{ij}$ is the weight in the lower-dimensional space transformed from the distance between points $i$ and $j$, and is defined as follows:
\[w'_{ij} = \left(1 + a \cdot d'(i,j)^{2b}\right)^{-1}\]
where $d'(i,j)$ is the distance between points $i$ and $j$ in the lower-dimensional space. The parameters $a$ and $b$ are derived from the $min\_dist$ and spread parameters.

\vspace{5mm}
\noindent 
\begin{tabular}{|L{8.5cm}|} 
	\hline
	\textbf{Algorithm 3:} Uniform manifold approximation and projection (UMAP) \\
	\hline
	\textbf{Input:} High-dimensional data $X$, dimensionality of the reduced space $n\_components$, number of neighbors $n\_neighbors$, minimum distance in the low-dimensional space $min\_dist$, number of optimization epochs $n\_iter$.
	\\
	\textbf{Process:} 
	\\ Initialize low-dimensional embeddings $Y$ to random values.
	\\ \textbf{For} each iteration $t=1$ \textbf{in} $n\_iter$:
	\\ \quad \textbf{For} each point $i$ \textbf{in} Data:
	\\ \quad \quad Calculate distance between point $i$ and all other \\ \quad \quad points.
	\\ \quad \quad Select the closest $n\_neighbors$ points as neighbors.
	\\ \quad \quad Calculate local scale $\sigma_i$ to balance Gaussian \\ \quad \quad distribution weights.
	\\ \quad \quad \textbf{For} each neighbor point $j$ \textbf{in} neighbor set:
	\\ \quad \quad \quad Calculate connectivity weight $w_{ij} = e^{-(d_{ij} / 2\sigma_i)^2}.$
	\\ \quad \quad \quad Calculate weights $w'_{ij}$ in low-dimensional space.
	\\ \quad \quad \quad Compute loss contribution and update the \\ \quad \quad \quad low-dimensional coordinates of point $i$.
	\\ \textbf{Output:} The low-dimensional embeddings $Y$.\\
	\hline
\end{tabular}
\vspace{5mm}

\section{Experimental results and discussions}
\subsection{Persistent Diagrams Analysis for Different Sleep Stages}

\begin{figure*}[!htb]  
	\centering           
	\includegraphics[width=0.8\textwidth]{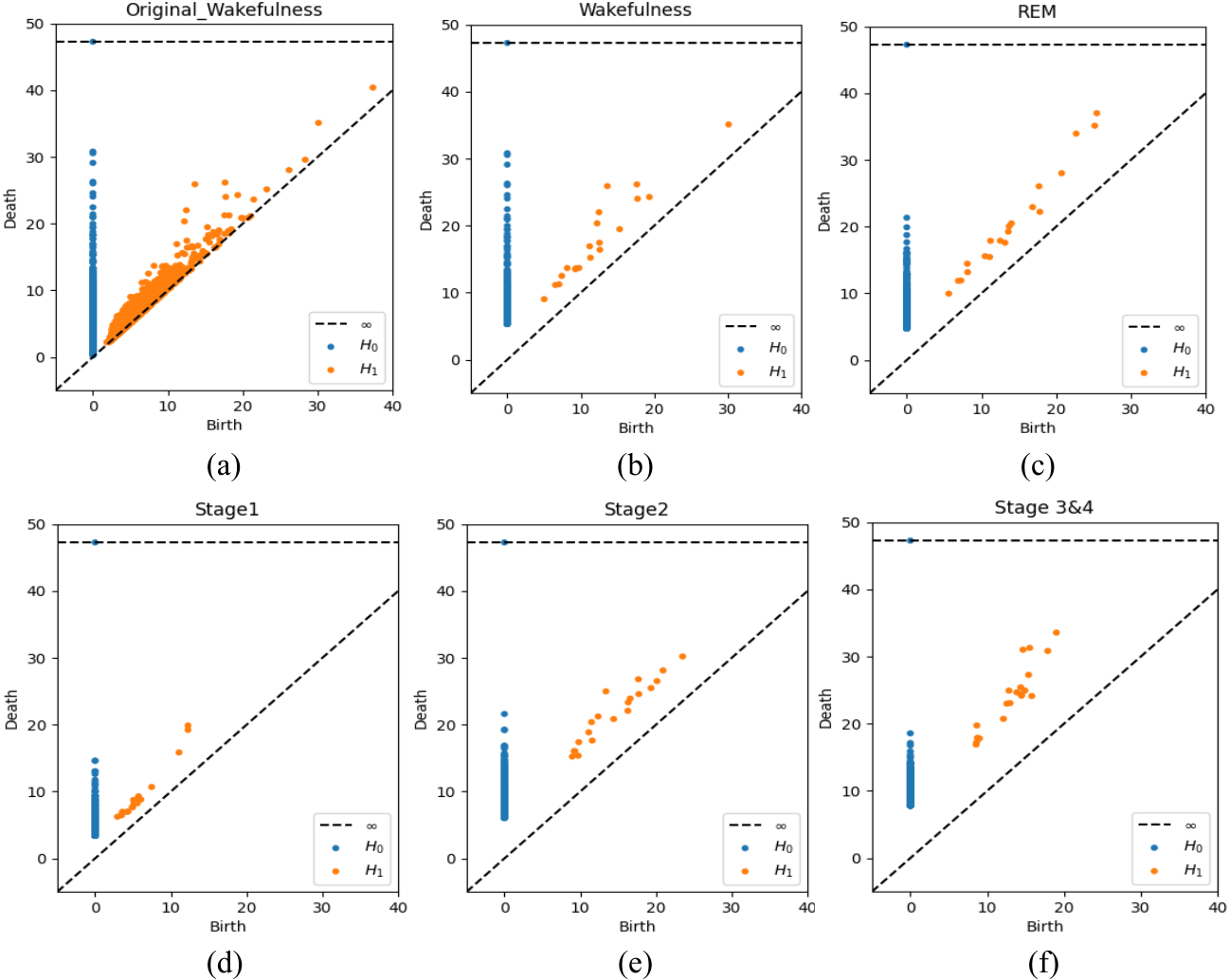}  
	\caption{Persistent diagrams for different sleep stages.}  
	\label{fig:routine}  
\end{figure*}

Persistent diagram is the visual representation of the persistent homology analysis, which can reflect the complexity of data structure. In physiology, different sleep stages exhibit varying levels of brain electrical activities, generating EEG signals of different complexities. Therefore, we conducted the persistent homology analysis on EEG signals from different sleep stages of the same individual in the Sleep-EDF dataset, with results presented in Fig. 6. Considering that EEG signals are high-dimensional time series, the resulting persistent intervals are quite cluttered, as shown in Fig. 6(a). As mentioned before, the intervals further from the $y=x$ line are more robust. Consequently, we applied a filter to the results, displaying only the top 500 points in $H_0$ and the top 20 points in $H_1$ that are furthest from the $y=x$ line. Figs. 6(b)-(f) are the filtered persistent diagrams for different sleep stages.

Since the EEG signal is a one-dimensional time series, direct application of Persistent Homology analysis hardly captures voids higher than 0-dimensional, failing to fully reflect the structural features of the data. Therefore, we employed Takens' embedding~\cite{takens2006detecting}, which can transform a time series into a higher-dimensional point cloud named embedding vector. In a time series $x_n$, given the time delay $ \tau \in \mathbb{N} $ and the embedding dimension $d \in \mathbb{N}$, the embedding vector is defined as $v_n = [x_n, x_{n+\tau}, x_{n+2\tau}, …, x_{n+(d-1)\tau}]$. Subsequently, persistent homology analysis will quantify the number of $0$-dimensional voids, denoted as $H_0$, and the number of $1$-dimensional voids, denoted as $H_1$, and generate the corresponding persistent diagrams.

We can observe that among the $H_0$ type points distributed along the $x=0$ line, the difference between the maximum $\epsilon_{\max}$ and the minimum $\epsilon_{\max}$ (the blue quasi-continuous segment composed of dense points) significantly decreases as sleep progresses from REM to Stage 1. With the maximum $\epsilon_{\max}$ values remaining almost constant, this indicates that some $H_0$ points have become more robust as their distances from the $y=x$ line increases. This process corresponds to the reduction in brain activities as sleep deepens, leading to a gradual decrease in the amplitude variation of EEG signals. Because the differences in signal values diminish, the lifespans of voids identified by persistent homology analysis are extended, positioning them further from the $y=x$ line. This physiological phenomenon can also be observed through the $H_1$ type points. As sleep progresses from wakefulness and REM to Stages 1-3, the $H_1$ type points move further away from the $y=x$ line. Additionally, we can observe that the dispersion of $H_1$ type points gradually decreases with deeper sleep stages. During wakefulness, brain activities are relatively active and complex. At this time, various parts of the brain interact extensively and dynamically, potentially forming a variety of loop structures and connection pathways, resulting in a higher dispersion of data points. As sleep deepens, entering the sleep stages, especially the deep sleep of Stages 3-4, the activity patterns of the brain become more synchronized and orderly. In these stages, the functional connections of the brain may reduce, showing more concentrated and stable loop structures, leading to a decrease in the dispersion of data points. This concentrated loop structure may be related to physiological processes such as cellular repair, memory consolidation, and metabolic waste clearance.

\subsection{Classification Performances for Dimensionality Reduction Algorithms}

\begin{table*}[]
	\caption{Classification Performances based on Physionet Dataset}
	\label{table_example}
	\centering
	\begin{tabular}{|c|c|ccc|ccccc|}
		\hline
		\multirow{2}{*}{Methods}  & \multirow{2}{*}{Features}                                                                       & \multicolumn{3}{c|}{Overall metrics}                                                                                                                                                                                                     & \multicolumn{5}{c|}{Per-class F1-score (F1)}                                                                                                                                                                                                                                                                                                                                                                      \\
		&                                                                                                 & ACC                                                                                 & MF1                                                                                & $\kappa$                                                             & W                                                                                  & N1                                                                                 & N2                                                                                 & N3                                                                                 & REM                                                           \\ \hline
		\multirow{3}{*}{Original} & Spectral-temporal features                                                                      & \multicolumn{1}{c|}{\begin{tabular}[c]{@{}c@{}}78.1\%$\pm$0.3\%\end{tabular}} & \multicolumn{1}{c|}{\begin{tabular}[c]{@{}c@{}}0.663$\pm$0.007\end{tabular}} & \begin{tabular}[c]{@{}c@{}}0.675$\pm$0.007\end{tabular} & \multicolumn{1}{c|}{\begin{tabular}[c]{@{}c@{}}0.844\end{tabular}} & \multicolumn{1}{c|}{\begin{tabular}[c]{@{}c@{}}\textbf{0.361}\end{tabular}} & \multicolumn{1}{c|}{\begin{tabular}[c]{@{}c@{}}0.808\end{tabular}} & \multicolumn{1}{c|}{\begin{tabular}[c]{@{}c@{}}0.635\end{tabular}} & \begin{tabular}[c]{@{}c@{}}0.663\end{tabular} \\ \cline{2-10} 
		& Topological features                                                                            & \multicolumn{1}{c|}{\begin{tabular}[c]{@{}c@{}}67.4\%$\pm$0.6\%\end{tabular}} & \multicolumn{1}{c|}{\begin{tabular}[c]{@{}c@{}}0.540$\pm$0.006\end{tabular}} & \begin{tabular}[c]{@{}c@{}}0.508$\pm$0.010\end{tabular} & \multicolumn{1}{c|}{\begin{tabular}[c]{@{}c@{}}0.717\end{tabular}} & \multicolumn{1}{c|}{\begin{tabular}[c]{@{}c@{}}0.245\end{tabular}} & \multicolumn{1}{c|}{\begin{tabular}[c]{@{}c@{}}0.705\end{tabular}} & \multicolumn{1}{c|}{\begin{tabular}[c]{@{}c@{}}0.634\end{tabular}} & \begin{tabular}[c]{@{}c@{}}0.415\end{tabular} \\ \cline{2-10} 
		& \begin{tabular}[c]{@{}c@{}}Spectral-temporal and topological features\end{tabular} & \multicolumn{1}{c|}{\begin{tabular}[c]{@{}c@{}}78.2\%$\pm$0.3\%\end{tabular}} & \multicolumn{1}{c|}{\begin{tabular}[c]{@{}c@{}}0.665$\pm$0.006\end{tabular}} & \begin{tabular}[c]{@{}c@{}}0.677$\pm$0.006\end{tabular} & \multicolumn{1}{c|}{\begin{tabular}[c]{@{}c@{}}0.845\end{tabular}} & \multicolumn{1}{c|}{\begin{tabular}[c]{@{}c@{}}0.360\end{tabular}} & \multicolumn{1}{c|}{\begin{tabular}[c]{@{}c@{}}0.809\end{tabular}} & \multicolumn{1}{c|}{\begin{tabular}[c]{@{}c@{}}\textbf{0.646}\end{tabular}} & \begin{tabular}[c]{@{}c@{}}0.663\end{tabular} \\ \hline
		
		\multirow{3}{*}{PCA}      & Spectral-temporal features                                                                      & \multicolumn{1}{c|}{\begin{tabular}[c]{@{}c@{}}73.1\%$\pm$0.5\%\end{tabular}} & \multicolumn{1}{c|}{\begin{tabular}[c]{@{}c@{}}0.568$\pm$0.008\end{tabular}} & \begin{tabular}[c]{@{}c@{}}0.593$\pm$0.009\end{tabular} & \multicolumn{1}{c|}{\begin{tabular}[c]{@{}c@{}}0.766\end{tabular}} & \multicolumn{1}{c|}{\begin{tabular}[c]{@{}c@{}}0.262\end{tabular}} & \multicolumn{1}{c|}{\begin{tabular}[c]{@{}c@{}}0.765\end{tabular}} & \multicolumn{1}{c|}{\begin{tabular}[c]{@{}c@{}}0.510\end{tabular}} & \begin{tabular}[c]{@{}c@{}}0.531\end{tabular} \\ \cline{2-10} 
		& Topological features                                                                            & \multicolumn{1}{c|}{\begin{tabular}[c]{@{}c@{}}65.7\%$\pm$0.6\%\end{tabular}} & \multicolumn{1}{c|}{\begin{tabular}[c]{@{}c@{}}0.468$\pm$0.010\end{tabular}} & \begin{tabular}[c]{@{}c@{}}0.462$\pm$0.014\end{tabular} & \multicolumn{1}{c|}{\begin{tabular}[c]{@{}c@{}}0.651\end{tabular}} & \multicolumn{1}{c|}{\begin{tabular}[c]{@{}c@{}}0.164\end{tabular}} & \multicolumn{1}{c|}{\begin{tabular}[c]{@{}c@{}}0.690\end{tabular}} & \multicolumn{1}{c|}{\begin{tabular}[c]{@{}c@{}}0.492\end{tabular}} & \begin{tabular}[c]{@{}c@{}}0.350\end{tabular} \\ \cline{2-10} 
		
		& \begin{tabular}[c]{@{}c@{}}Spectral-temporal and topological features\end{tabular} & \multicolumn{1}{c|}{\begin{tabular}[c]{@{}c@{}}73.2\%$\pm$0.5\%\end{tabular}} & \multicolumn{1}{c|}{\begin{tabular}[c]{@{}c@{}}0.569$\pm$0.008\end{tabular}} & \begin{tabular}[c]{@{}c@{}}0.596$\pm$0.010\end{tabular} & \multicolumn{1}{c|}{\begin{tabular}[c]{@{}c@{}}0.768\end{tabular}} & \multicolumn{1}{c|}{\begin{tabular}[c]{@{}c@{}}0.263\end{tabular}} & \multicolumn{1}{c|}{\begin{tabular}[c]{@{}c@{}}0.763\end{tabular}} & \multicolumn{1}{c|}{\begin{tabular}[c]{@{}c@{}}0.513\end{tabular}} & \begin{tabular}[c]{@{}c@{}}0.528\end{tabular} \\ \hline
		
		\multirow{3}{*}{t-SNE}    & Spectral-temporal features                                                                      & \multicolumn{1}{c|}{\begin{tabular}[c]{@{}c@{}}79.7\%$\pm$0.3\%\end{tabular}} & \multicolumn{1}{c|}{\begin{tabular}[c]{@{}c@{}}0.665$\pm$0.007\end{tabular}} & \begin{tabular}[c]{@{}c@{}}0.697$\pm$0.006\end{tabular} & \multicolumn{1}{c|}{\begin{tabular}[c]{@{}c@{}}\textbf{0.859}\end{tabular}} & \multicolumn{1}{c|}{\begin{tabular}[c]{@{}c@{}}0.356\end{tabular}} & \multicolumn{1}{c|}{\begin{tabular}[c]{@{}c@{}}\textbf{0.822}\end{tabular}} & \multicolumn{1}{c|}{\begin{tabular}[c]{@{}c@{}}0.611\end{tabular}} & \begin{tabular}[c]{@{}c@{}}\textbf{0.671}\end{tabular} \\ \cline{2-10} 
		
		& Topological features                                                                            & \multicolumn{1}{c|}{\begin{tabular}[c]{@{}c@{}}68.8\%$\pm$0.5\%\end{tabular}} & \multicolumn{1}{c|}{\begin{tabular}[c]{@{}c@{}}0.526$\pm$0.008\end{tabular}} & \begin{tabular}[c]{@{}c@{}}0.522$\pm$0.001\end{tabular} & \multicolumn{1}{c|}{\begin{tabular}[c]{@{}c@{}}0.722\end{tabular}} & \multicolumn{1}{c|}{\begin{tabular}[c]{@{}c@{}}0.210\end{tabular}} & \multicolumn{1}{c|}{\begin{tabular}[c]{@{}c@{}}0.719\end{tabular}} & \multicolumn{1}{c|}{\begin{tabular}[c]{@{}c@{}}0.568\end{tabular}} & \begin{tabular}[c]{@{}c@{}}0.416\end{tabular} \\ \cline{2-10} 
		
		& \begin{tabular}[c]{@{}c@{}}Spectral-temporal and topological features\end{tabular} & \multicolumn{1}{c|}{\begin{tabular}[c]{@{}c@{}}\textbf{79.8\%}$\pm$0.3\%\end{tabular}} & \multicolumn{1}{c|}{\begin{tabular}[c]{@{}c@{}}\textbf{0.666}$\pm$0.007\end{tabular}} & \begin{tabular}[c]{@{}c@{}}\textbf{0.698}$\pm$0.005\end{tabular} & \multicolumn{1}{c|}{\begin{tabular}[c]{@{}c@{}}0.859\end{tabular}} & \multicolumn{1}{c|}{\begin{tabular}[c]{@{}c@{}}0.355\end{tabular}} & \multicolumn{1}{c|}{\begin{tabular}[c]{@{}c@{}}0.822\end{tabular}} & \multicolumn{1}{c|}{\begin{tabular}[c]{@{}c@{}}0.621\end{tabular}} & \begin{tabular}[c]{@{}c@{}}0.670\end{tabular} \\ \hline
		
		\multirow{3}{*}{UMAP}     & Spectral-temporal features                                                                      & \multicolumn{1}{c|}{\begin{tabular}[c]{@{}c@{}}78.6\%$\pm$0.3\%\end{tabular}} & \multicolumn{1}{c|}{\begin{tabular}[c]{@{}c@{}}0.643$\pm$0.008\end{tabular}} & \begin{tabular}[c]{@{}c@{}}0.678$\pm$0.006\end{tabular} & \multicolumn{1}{c|}{\begin{tabular}[c]{@{}c@{}}0.841\end{tabular}} & \multicolumn{1}{c|}{\begin{tabular}[c]{@{}c@{}}0.320\end{tabular}} & \multicolumn{1}{c|}{\begin{tabular}[c]{@{}c@{}}0.815\end{tabular}} & \multicolumn{1}{c|}{\begin{tabular}[c]{@{}c@{}}0.591\end{tabular}} & \begin{tabular}[c]{@{}c@{}}0.644\end{tabular} \\ \cline{2-10} 
		
		& Topological features                                                                            & \multicolumn{1}{c|}{\begin{tabular}[c]{@{}c@{}}68.1\%$\pm$0.5\%\end{tabular}} & \multicolumn{1}{c|}{\begin{tabular}[c]{@{}c@{}}0.510$\pm$0.007\end{tabular}} & \begin{tabular}[c]{@{}c@{}}0.507$\pm$0.010\end{tabular} & \multicolumn{1}{c|}{\begin{tabular}[c]{@{}c@{}}0.701\end{tabular}} & \multicolumn{1}{c|}{\begin{tabular}[c]{@{}c@{}}0.195\end{tabular}} & \multicolumn{1}{c|}{\begin{tabular}[c]{@{}c@{}}0.712\end{tabular}} & \multicolumn{1}{c|}{\begin{tabular}[c]{@{}c@{}}0.557\end{tabular}} & \begin{tabular}[c]{@{}c@{}}0.391\end{tabular} \\ \cline{2-10} 
		
		& \begin{tabular}[c]{@{}c@{}}Spectral-temporal and topological features\end{tabular} & \multicolumn{1}{c|}{\begin{tabular}[c]{@{}c@{}}78.8\%$\pm$0.3\%\end{tabular}} & \multicolumn{1}{c|}{\begin{tabular}[c]{@{}c@{}}0.644$\pm$0.007\end{tabular}} & \begin{tabular}[c]{@{}c@{}}0.679$\pm$0.006\end{tabular} & \multicolumn{1}{c|}{\begin{tabular}[c]{@{}c@{}}0.841\end{tabular}} & \multicolumn{1}{c|}{\begin{tabular}[c]{@{}c@{}}0.321\end{tabular}} & \multicolumn{1}{c|}{\begin{tabular}[c]{@{}c@{}}0.814\end{tabular}} & \multicolumn{1}{c|}{\begin{tabular}[c]{@{}c@{}}0.596\end{tabular}} & \begin{tabular}[c]{@{}c@{}}0.646\end{tabular} \\ \hline
	\end{tabular}
\end{table*}

Following the routine in Fig. 5, the following metrics were calculated to evaluate different dimensionality reduction algorithms and the input features: Average Overall Accuracy (ACC), Macro-F1 Score (MF1), Cohen’s Kappa ($\kappa$), and the per-class F1-Score for each distinct sleep stage. These metrics were calculated after feeding the results of the dimensionality reduction algorithms into a K-Nearest Neighbors (KNN) classifier. As shown in Table II, for data of each subject, we calculated the mean and variance for the overall metrics, including ACC, MF1, and $\kappa$, as well as the mean for the per-class F1-Score. Regarding the input features, three different sets—spectral-temporal features, topological features, and spectral-temporal features—were input into the dimensionality reduction algorithms. In terms of dimensionality reduction algorithms, three of the most representative algorithms, PCA, t-SNE, and UMAP, were used to reduce the dimensionality of the input features. The 'Original' group represents the input features directly fed into the KNN classifier, serving as a reference group.

Previous analyses of persistent diagrams have fully demonstrated that the results of persistent homology analysis on EEG signals are informative and effective. This view is also corroborated in detail in Table II. Under the premise of a given dimensionality reduction algorithm, using spectral-temporal and topological features improved the overall metrics compared to using only spectral-temporal features. This phenomenon is also valid for most per-class F1-Scores. However, considering that the improvement is not very significant and that using topological features alone is not as effective as using spectral-temporal features, we believe that topological features serve as a powerful complement to traditional spectral-temporal features rather than a complete supplement. This also aligns with theoretical reasoning: persistent homology focuses more on extracting structural information from the data rather than the fluctuations within the data itself, whereas traditional spectral-temporal features do the opposite. Therefore, these two types of features effectively complement each other.

In the perspective of the dimensionality reduction algorithms, among the three, PCA was the only dimensionality reduction algorithm with all metrics lower than those of 'Original'. Considering that the locally optimal feature combinations obtained after using the RFECV algorithm probably still contain residual redundant features, this indicates that PCA is less effective at simultaneously removing redundant information and preserving useful information compared to manifold-based algorithms, t-SNE and UMAP. However, its simplicity allows it to be computationally efficient. Among the manifold-based dimensionality reduction algorithms, t-SNE performed better overall than UMAP. The highest ACC of 79.8\%, the highest MF1 of 0.666, and the highest $\kappa$ of 0.698 were all achieved when using spectral-temporal and topological features as the input data for t-SNE. However, t-SNE has more hyperparameters. Thus, it is computationally demanding and requires longer time to diverge to the optimal values of the hyperparameters. In contrast, UMAP, with almost comparable overall metrics to t-SNE, has fewer hyperparameter combinations and generally faster computational speed. Therefore, considering the overall algorithm performance and computational resource consumption, UMAP is the most optimal one among three.

\subsection{Visualization Performances for Dimensionality Reduction Algorithms}
\begin{figure*}[!htb]  
	\centering           
	\includegraphics[width=\textwidth]{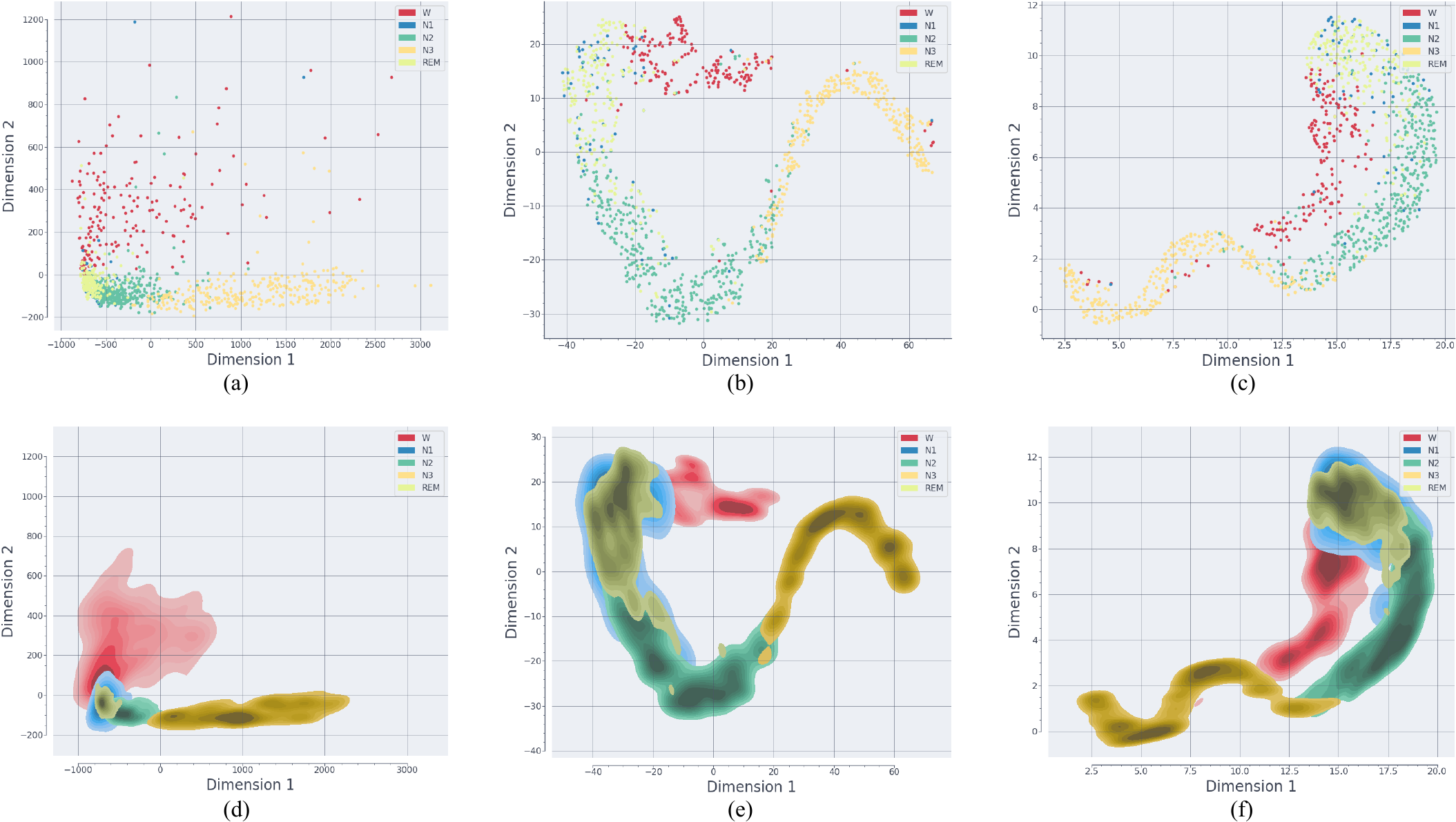}  
	\caption{Visualization performances of three kinds of dimensionality algorithms: PCA, t-SNE, and UMAP.}  
	\label{fig:routine}  
\end{figure*}

Fig. 7 shows the scatter plots of a subject's data from the Physionet dataset after dimensionality reduction to two dimensions using PCA, t-SNE, and UMAP, following the routine in Fig. 5. Figs. 7(a)-(c) correspond to the scatter plots for PCA, t-SNE, and UMAP, respectively. Figs. 7(d)-(f) are the Kernel Density Estimation (KDE) plots corresponding to Figs. 7(a)-(c). 

In Fig. 7(a), PCA demonstrates its capability to distinguish different sleep stages effectively. However, in Fig. 7(d), there is a significant overlap among the regions of stages N1, N2, and REM. This indicates that the distinctions between different sleep stages are not sufficiently clear, leading to potential misclassification by distance-based classification algorithms such as KNN. This is also corroborated by the overall metrics of PCA being lower than those of manifold-based algorithms. Figs. 7(b) and (c) show that manifold-based algorithms effectively cluster different sleep stages and achieve higher differentiation between them. However, in Figs. 7(e) and (f), the distinction between Stage N1 and REM is not clear, with some overlap between these stages. This is likely because the EEG signals for these two sleep stages are characterized by low amplitude and high frequency, affecting the dimensionality reduction outcomes. Nevertheless, from the overall classification results, t-SNE and UMAP demonstrate the potential of manifold-based algorithms for exploring more complex sleep patterns and other physiological phenomena using EEG signals.

\section{Conclusion}
Explainable dimensionality reduction algorithms are crucial for data analysis in fundamental scientific research, such as the exploration of sleep patterns. However, their performance is always inferior to some advanced machine learning algorithms such as deep learning. We combined the RFECV algorithm with explainable dimensionality reduction algorithms to construct a two-stage hierarchical and explainable feature selection framework. This framework successfully improved the performance of dimensionality reduction algorithms on high-dimensional EEG data. Considering that the performance of the proposed framework may be sensitive to input data, we performed a visualization analysis on the EEG data from different sleep stages through persistent homology and calculated topological features to enrich traditional spectral-temporal features. Based on 51 input features, including spectral-temporal and topological features, we conducted comparative experiments on three different dimensionality reduction algorithms: PCA, t-SNE and UMAP. Ultimately, the t-SNE algorithm performed the best, achieving an accuracy of 79.8\%. Overall, the proposed two-stage hierarchical framework enhances the performance of explainable dimensionality reduction algorithms and benefits the exploration of sleep stages and their data visualization.

\bibliographystyle{IEEEtran}
\bibliography{ref}

\end{document}